\documentclass{article}

\usepackage{arxiv}
\usepackage{times}
\usepackage{booktabs}
\usepackage{graphicx}
\usepackage{amsmath}
\usepackage{amssymb}
\usepackage{cite}
\usepackage{balance}
\usepackage[table]{xcolor}
\usepackage{subcaption}
\usepackage{tikz}
\usetikzlibrary{shapes.geometric, arrows, positioning}

\usepackage[pagebackref=true,breaklinks=true,colorlinks,bookmarks=false]{hyperref}

\begin{document}

\title{Investigation of accuracy and bias in face recognition trained with synthetic data}

\author{
Pavel Korshunov$^{1}$,
Ketan Kotwal$^{1}$,
Christophe Ecabert$^{1}$,\\
{\bf Vidit Vidit$^{1}$,
Amir Mohammadi$^{1}$, and
S\'{e}bastien Marcel$^{1,2}$}\\
$^{1}$~Idiap Research Institute, Martigny, Switzerland\\
$^{2}$~Universit\'{e} de Lausanne (UNIL), Lausanne, Switzerland\\
{\tt\small \{firstname.lastname\}@idiap.ch}
}

\maketitle
\thispagestyle{empty}

\begin{abstract}
Synthetic data has emerged as a promising alternative for training face recognition (FR) models, offering advantages in scalability, privacy compliance, and potential for bias mitigation. However, critical questions remain on whether both high accuracy and fairness can be achieved with synthetic data. In this work, we evaluate the impact of synthetic data on bias and performance of FR systems. We generate balanced face dataset, FairFaceGen, using two state of the art text-to-image generators, Flux.1-dev and Stable Diffusion v3.5 (SD35), and combine them with several identity augmentation methods, including Arc2Face and four IP-Adapters. By maintaining equal identity count across synthetic and real datasets, we ensure fair comparisons when evaluating FR performance on standard (LFW, AgeDB-30, etc.) and challenging IJB-B/C benchmarks and FR bias on Racial Faces in-the-Wild (RFW) dataset. Our results demonstrate that although synthetic data still lags behind the real datasets in the generalization on IJB-B/C, demographically balanced synthetic datasets, especially those generated with SD35, show potential for bias mitigation.
We also observe that the number and quality of intra-class augmentations significantly affect FR accuracy and fairness. These findings provide practical guidelines for constructing fairer FR systems using synthetic data.
\end{abstract}

\section{Introduction}
\label{sec:intro}

The use of synthetic data to train face recognition (FR) models has gained increasing attention in recent years, primarily, due to its potential to avoid ethical, legal, and licensing challenges associated with using real facial images, especially in the context of privacy regulations such as the GDPR~\cite{melzi2023gandiffface}.  Synthetic data offers the potential to generate large-scale datasets to train models for commercial use without infringing on individual privacy, hence, facilitating the development of safer FR systems. Moreover, it allows a more fine-grained control over the data generation, which potentially can help mitigating biases in FR systems.

\begin{figure}[h]
\setlength\tabcolsep{1pt} 
\centering
\begin{subfigure}{0.25\columnwidth}
\includegraphics[width=\linewidth]{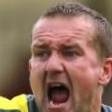}
	\caption{WebFace42M}
	\label{fig:webface}
\end{subfigure}
\qquad
\begin{subfigure}{0.25\columnwidth}
\includegraphics[width=\linewidth]{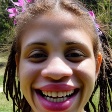}
	\caption{Digi2Real}
	\label{fig:digi2real}
\end{subfigure}
\qquad
\begin{subfigure}{0.25\columnwidth}
\includegraphics[width=\linewidth]{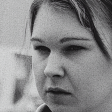}
	\caption{HyperFace}
	\label{fig:hyperface}
\end{subfigure}\\
\vspace{2mm}
\begin{subfigure}{0.25\columnwidth}
\includegraphics[width=\linewidth]{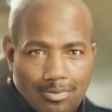}
	\caption{DCFace}
	\label{fig:dcface}
\end{subfigure}
\qquad
\begin{subfigure}{0.25\columnwidth}
\includegraphics[width=\linewidth]{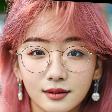}
	\caption{Stable Diffusion v3.5}
	\label{fig:sd35example}
\end{subfigure}
\qquad
\begin{subfigure}{0.25\columnwidth}
\includegraphics[width=\linewidth]{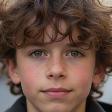}
	\caption{Flux.1-dev}
	\label{fig:fluxexample}
\end{subfigure}

\caption{Examples of images cropped to $112 \times 112$ from WebFace42M~\cite{Zhu2021WebFace260M} and synthetic datasets Digi2Real~\cite{George2025Digi2Real}, HyperFace~\cite{Otroshi2024HyperFace}, DCFace~\cite{Kim2023DCFace}, and generated by us with SD35 and Flux.}
\label{fig:basedata}
\end{figure}

The main focus of the recent work is on generating synthetic face datasets that could be used to train FR models with performance approaching that of models trained on real data~\cite{melzi2023gandiffface, Fadi2023IDiffFace, Kim2023DCFace, geissbuhler2024synthetic, Otroshi2024HyperFace, George2025Digi2Real, yeung2025variface}. However, several issues are still missing from the current research discourse pertaining to training FR models with synthetic data:

\begin{itemize}
    \item \textbf{Dual-generator framework:} Synthetic face data generation often employs a two-stage process: a \emph{seed generator} for creation of distinct identities and an \emph{augmentation generator} for producing intra-class variations such as different poses, lighting conditions, and expressions. Although considerable effort is directed toward enhancing the diversity of seed identities~\cite{George2025Digi2Real, Otroshi2024HyperFace}, the role of augmentation generators in influencing FR performance remains underexplored. 

    \item \textbf{Unfair dataset comparisons:} Comparative studies between synthetic and real datasets frequently suffer from inconsistencies in dataset sizes and compositions. A synthetic dataset with $10\,\mathrm{K}$ identities and $64$ images per identity gets compared with another dataset of $50\,\mathrm{k}$ identities and $20$ images per identity or to a WebFace-12M's $1.5\,\mathrm{M}$ identities and $12\,\mathrm{M}$ images~\cite{Zhu2021WebFace260M}. In these comparisons, the number of identities are different and the total number of images are different, which are the parameters affecting the performance of the model. Such mismatches conceal the true effect of generation parameters~\cite{atzori2024impact} and a fair evaluation needs to acknowledge the issue and equalize either (i) the number of identities and total images or (ii) the images-per-identity ratio. 
    
    \item \textbf{Dependence on real data:} Most synthetic face generation methods, including generative adversarial networks (GANs) and diffusion-based models, are trained on datasets of real facial images~\cite{Korshunov2023deepav}. Many generative methods also rely on an existing FR models pretrained on large datasets of real faces. Therefore, the generated synthetic data inherently reflects the biases present in the original datasets \cite{huber2024bias}. This dependency questions the purity of synthetic datasets and their efficacy in mitigating bias.

    \item \textbf{Lack of bias evaluation:} Synthetic data generation allows, at least in principle, for controlled manipulation of demographic attributes (e.g., race, gender, age), enabling balanced training data that may improve fairness~\cite{Qraitem2024FTR,kotwal2025review}. However, comprehensive evaluations of bias in FR models trained solely on synthetic data are scarce. Existing studies suggest that synthetic data can mirror the biases of their training datasets, emphasizing the need for deliberate balancing strategies~\cite{huber2024bias}.
\end{itemize}

\begin{figure}[htb]
\centering
\includegraphics[width=0.45\columnwidth]{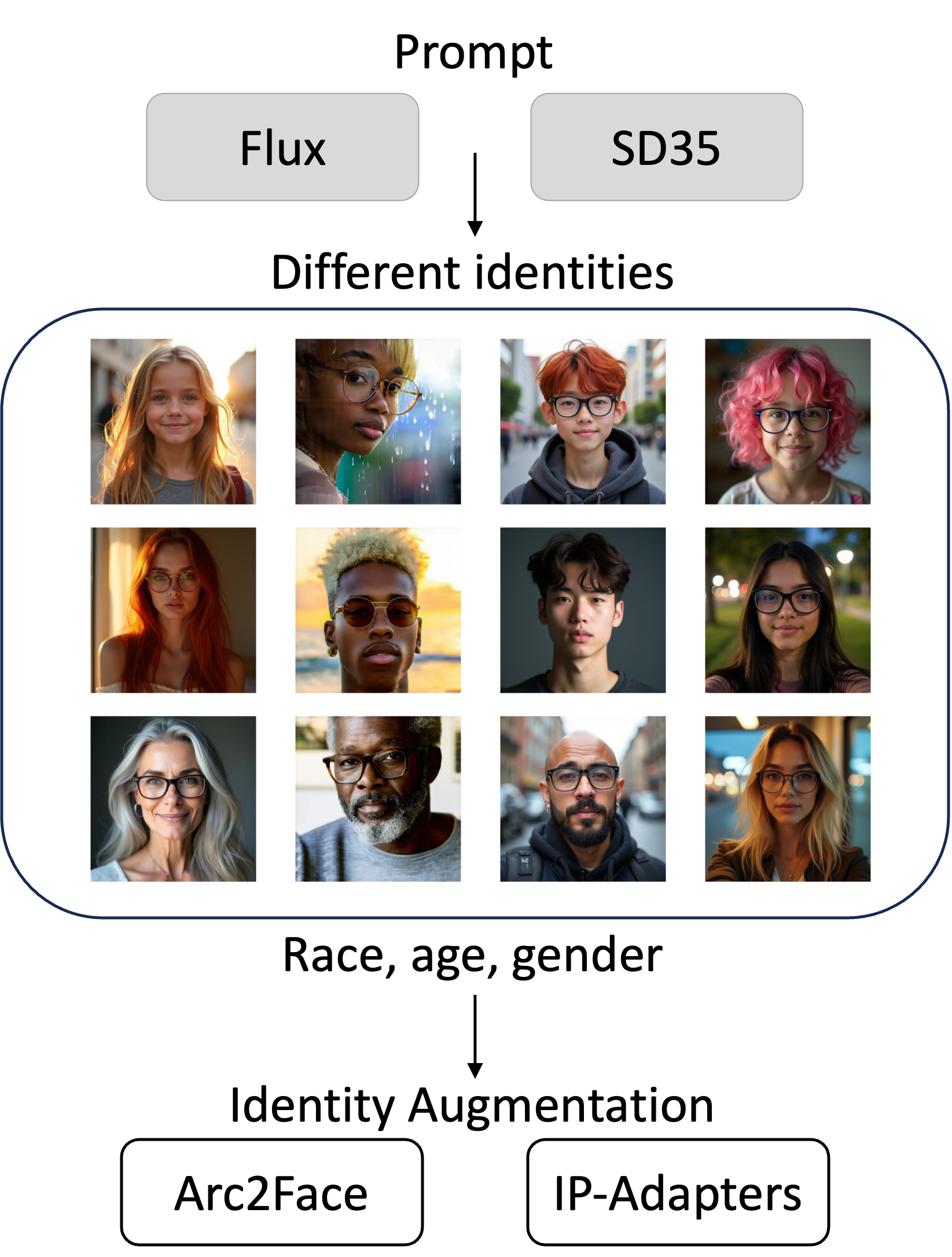}
\caption{Generation process of \textit{FairFaceGen} dataset.}
\label{fig:diagram}
\end{figure}

 In this work, we aim to systematically investigate the impact of synthetic data generation on the performance and bias of FR models. Specifically, we:

\begin{enumerate}
    \item Evaluate two state-of-the-art text-to-image models, Flux.1-dev~\cite{yang2024flux}, referred to as Flux in this paper, and Stable Diffusion~v3.5~\cite{stabilityai2024sd3}, referred to as SD35, for generation of different identities to investigate the impact of the \emph{seed generators} on the resulting FR model.  	

    \item Use prompt-based techniques with Flux and SD35 to generate balanced synthetic dataset, \textit{FairFaceGen}\footnote{ \url{https://www.idiap.ch/paper/fairfacegen}}  (see Figure~\ref{fig:diagram} for the illustration), across gender, age (categorized into seven groups from children to seniors), and race (White, Black, Asian, and Hispanic). This approach to generating balanced data does not require special model training or tuning, which, if effective, makes it very practical and simple.

    \item Compare several augmentation generators, including Arc2Face~\cite{Paraperas2024Arc2Face} pre-trained on WebFace-42M and four face IP-Adapter variants~\cite{Zhang2023IPAdapter}, with Stable Diffusion v1.5 (SD15) or Stable Diffusion XL (SDXL) models relying on the embeddings from either InsightFace~\cite{Deng2020CVPR} or CLIP~\cite{radford2021clip}, pretrained on LAION-2B~\cite{schuhmann2022laion} and COYO-700M~\cite{kakaobrain2022coyo} datasets.

    \item Measure inter-identity and intra-identity similarity with an independent EdgeFace~\cite{george2024edgeface} face recognition model and compare different variations of the synthetic data with HyperFace~\cite{Otroshi2024HyperFace}, Digi2Real~\cite{George2025Digi2Real} and a size-matched 10K identities subset from the WebFace-42M, which is a cleaned version of WebFace-260M~\cite{Zhu2021WebFace260M}.
    
    \item Besides using standard benchmarking datasets, including LFW~\cite{huang2008labeled} with its variants, AgeDB-30~\cite{moschoglou2017agedb} and others, we report verification performance on the challenging IJB-B/C~\cite{Whitelam2017IJBB, Maze2018IJBC} and bias on Racial Faces in-the-Wild (RFW)~\cite{Wang2019RFW} datasets, comparing our balanced sets to synthetic datasets, DCFace~\cite{Kim2023DCFace}, Digi2Real~\cite{George2025Digi2Real}, HyperFace~\cite{Otroshi2024HyperFace}, and to real datasets, CASIA-WebFace~\cite{yi2014learning} and a 10K identities subset of WebFace 42M~\cite{Zhu2021WebFace260M}.
\end{enumerate}
 
In this paper, our objectives are i) to provide quantitative guidelines on choosing \emph{seed} and \emph{augmentation generators}, ii) to investigate the impact of the number of images per identity on the FR model performance, and iii) to  evaluate how the demographically balanced synthetic datasets mitigate bias. 

\newcolumntype{M}[1]{>{\centering\arraybackslash}m{#1}} 

 \begin{figure*}[htb]
\setlength\tabcolsep{1pt} 
\centering
\begin{tabular}{@{} M{0.16\linewidth} M{0.16\linewidth} M{0.16\linewidth} M{0.16\linewidth} M{0.16\linewidth} M{0.16\linewidth} @{}}

\includegraphics[width=0.16\textwidth]{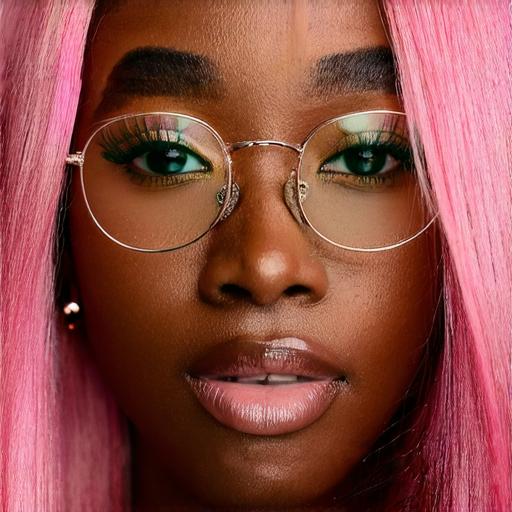}
& \includegraphics[width=0.16\textwidth]{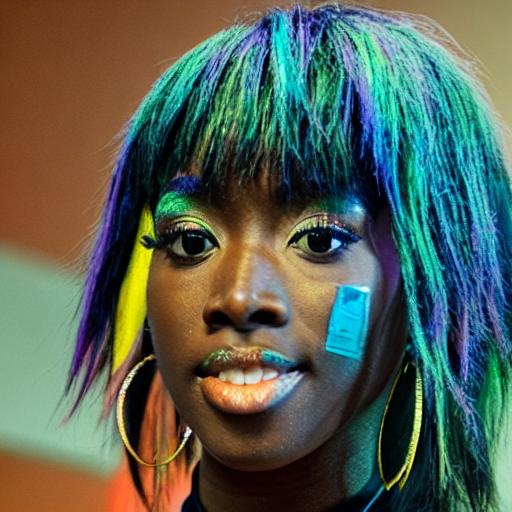}
& \includegraphics[width=0.16\textwidth]{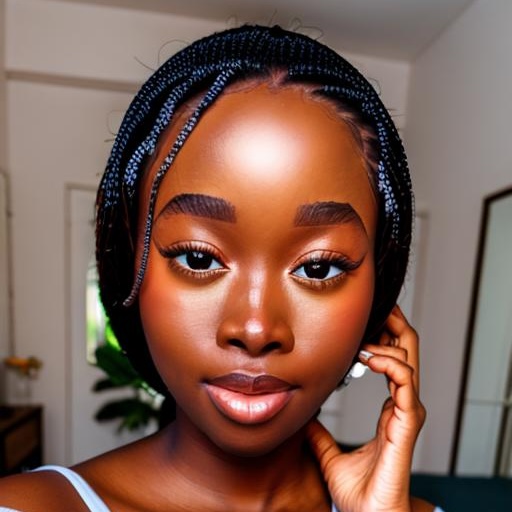}
& \includegraphics[width=0.16\textwidth]{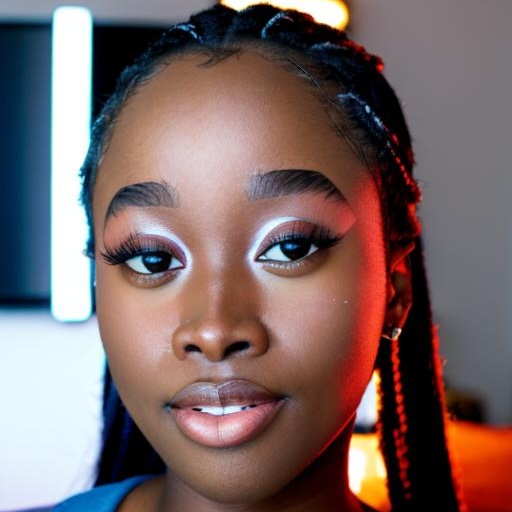}
& \includegraphics[width=0.16\textwidth]{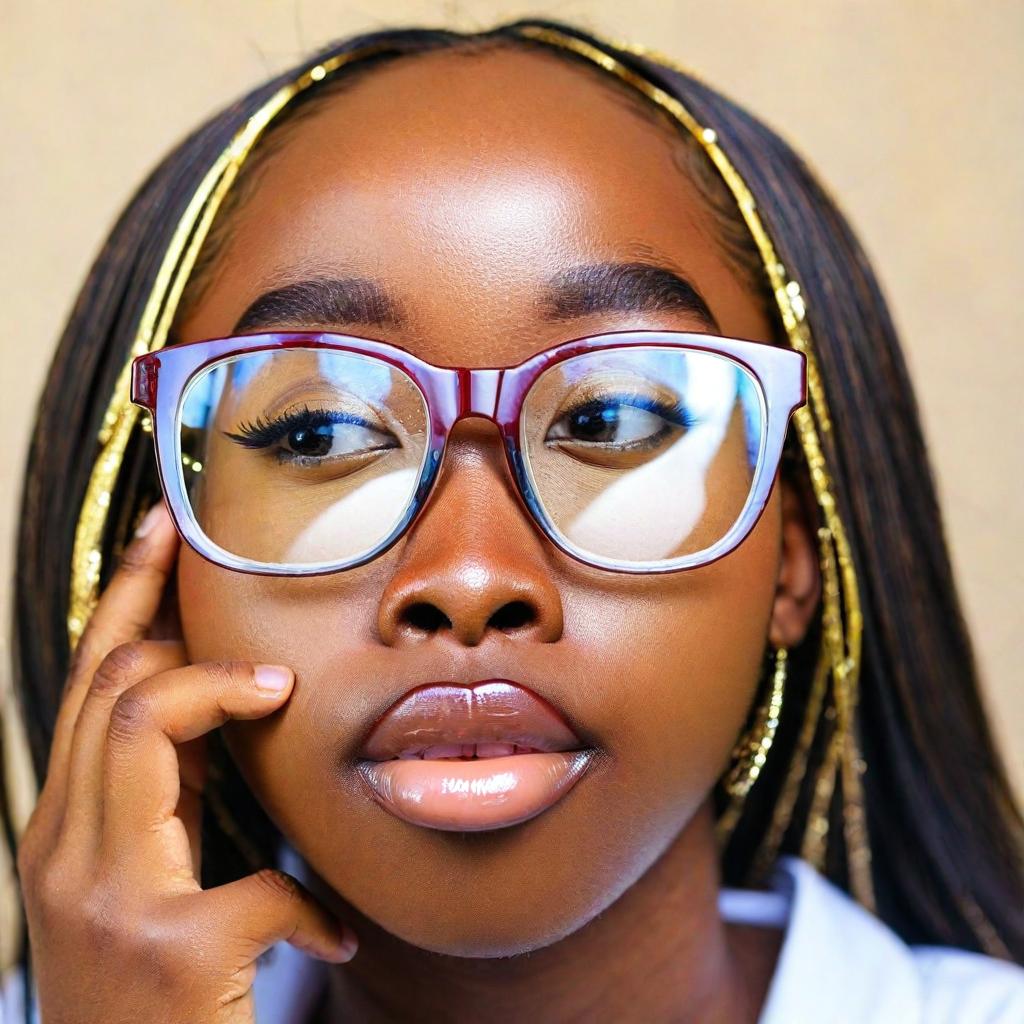}
& \includegraphics[width=0.16\textwidth]{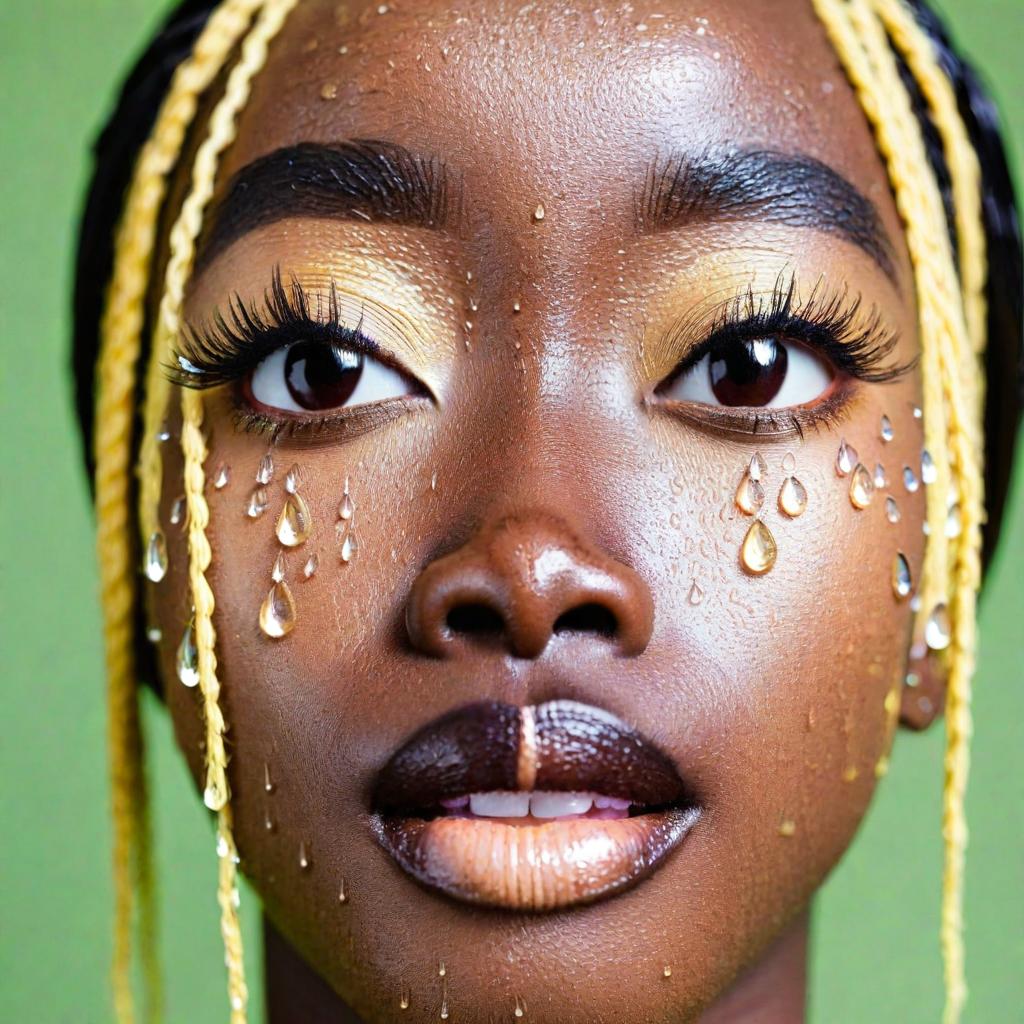}
\\

\includegraphics[width=0.16\textwidth]{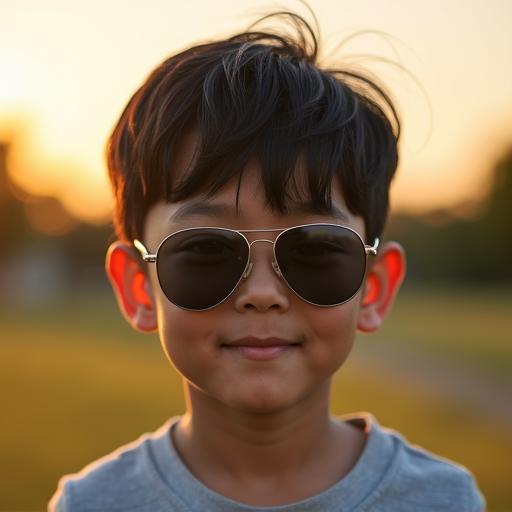}
& \includegraphics[width=0.16\textwidth]{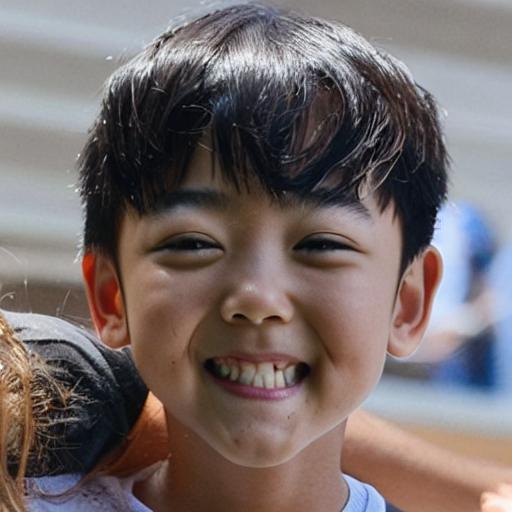}
& \includegraphics[width=0.16\textwidth]{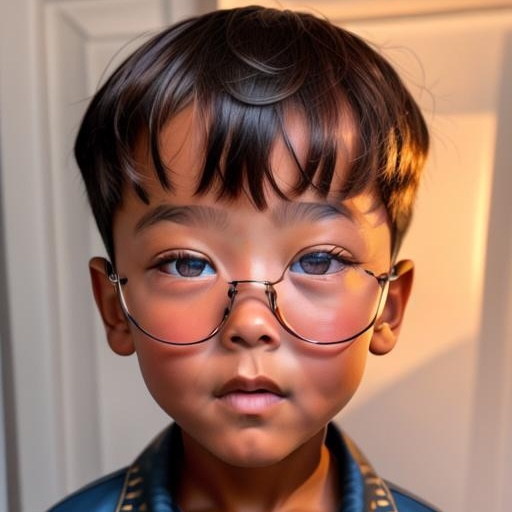}
& \includegraphics[width=0.16\textwidth]{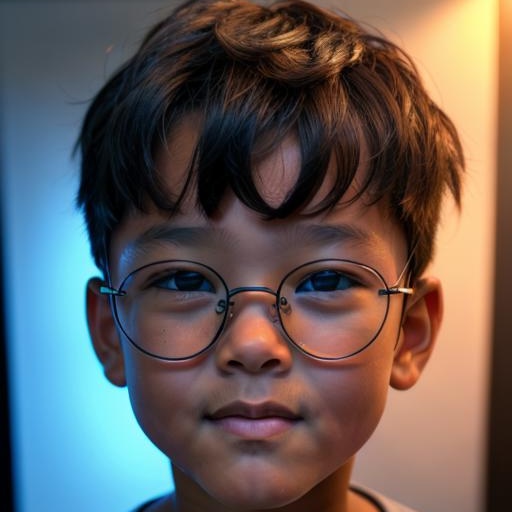}
& \includegraphics[width=0.16\textwidth]{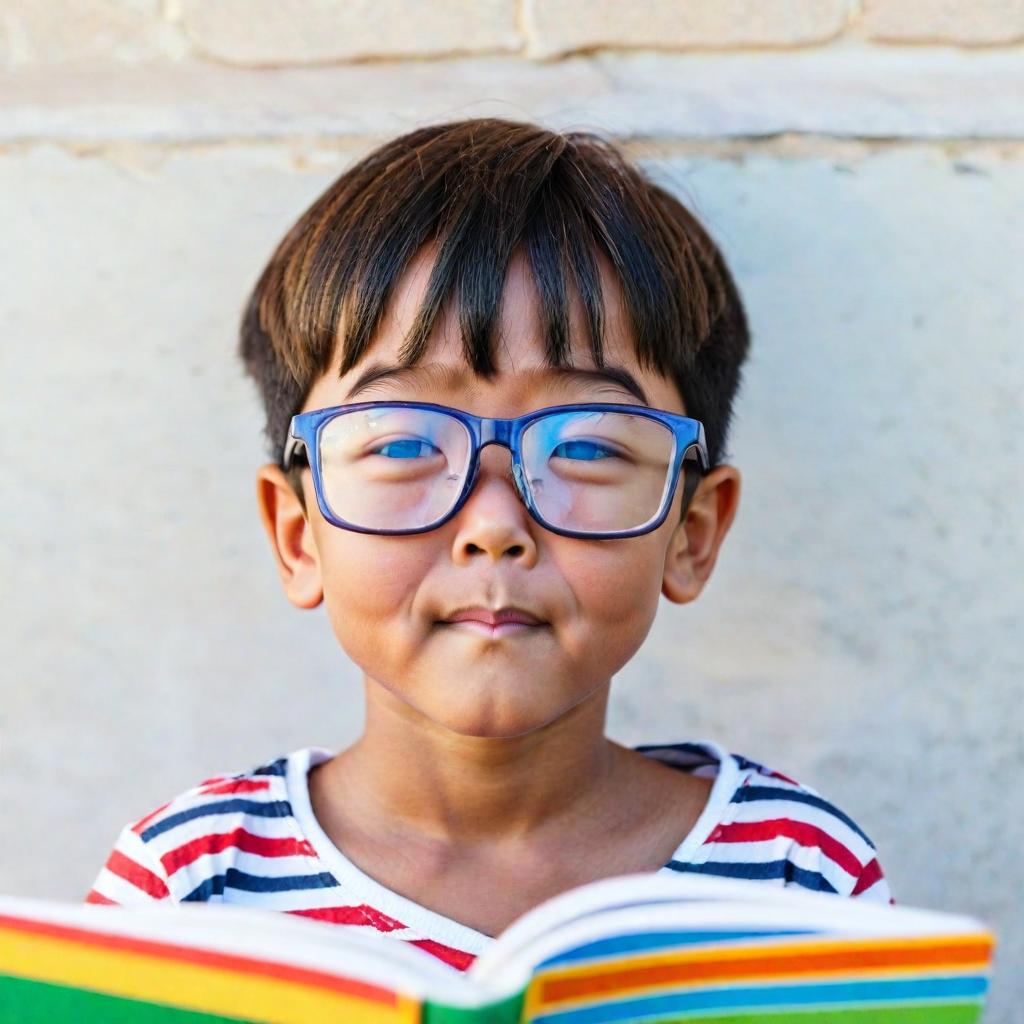}
& \includegraphics[width=0.16\textwidth]{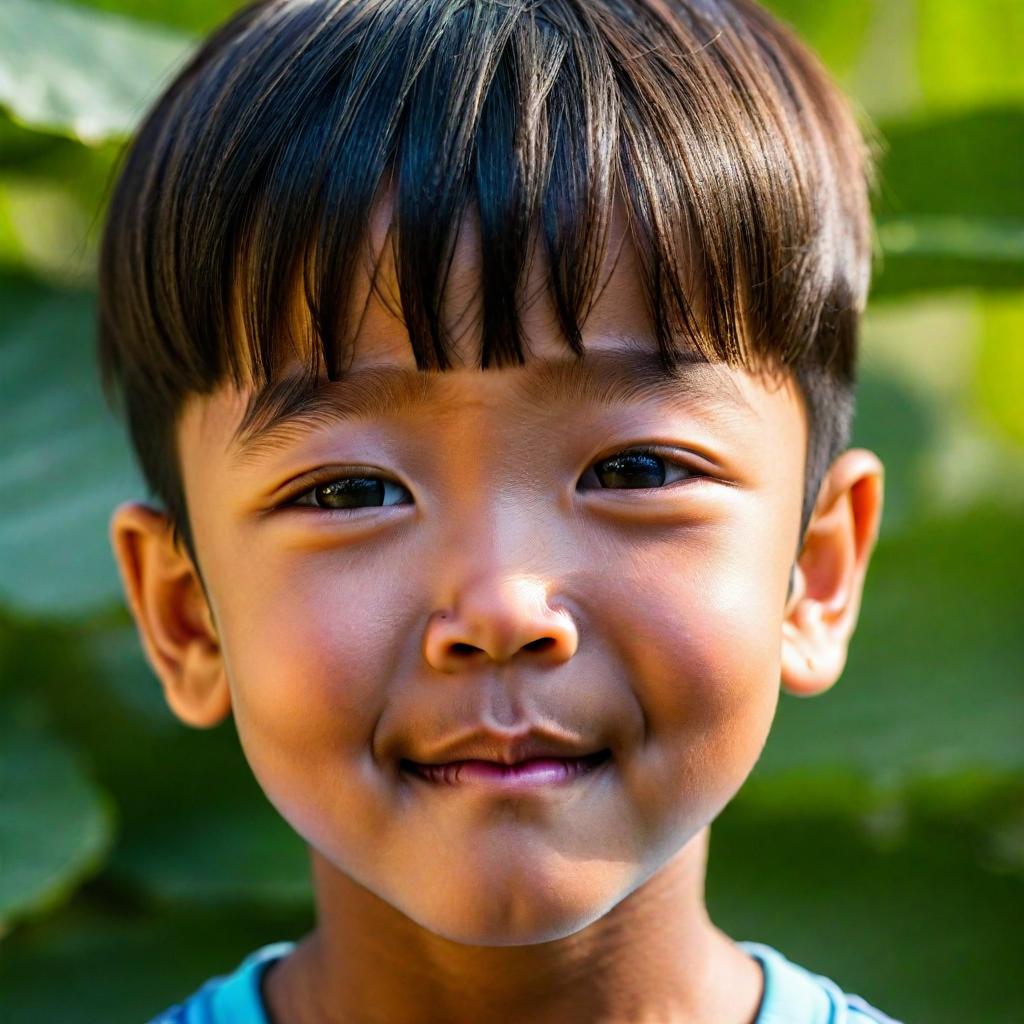}
\\
\subcaption{Seed SD35/Flux}\label{subfig:a} & \subcaption{Arc2Face}\label{subfig:b} & \subcaption{SD15 FaceID}\label{subfig:c} & \subcaption{SD15 CLIP}\label{subfig:d} & \subcaption{SDXL FaceID}\label{subfig:e} & \subcaption{SDXL CLIP}\label{subfig:f}
\end{tabular}

\caption{Seed generators SD35 (top of (a)) and Flux (bottom of (a)) and their augmentations generated with Arc2Face and IP-Adapters.}
\label{fig:flux}
\end{figure*}

\section{Related work}
\label{sec:related}

\paragraph{Synthetic identities for face recognition}
Early work showed that high-fidelity GANs such as StyleGAN2~\cite{Karras2020StyleGAN2}
can support FR pre-training provided sufficient inter-
and intra-class diversity.
Diffusion-based methods (e.g.\ DCFace~\cite{Kim2023DCFace}) improved identity
preservation, but these and subsequent works~\cite{George2025Digi2Real, yeung2025variface,Otroshi2024HyperFace} still combine \emph{seed} identity creation
and \emph{augmentation} of each identity. The impact of the latter stage on FR
accuracy remains under-explored.

\paragraph{Augmentation generators and identity fidelity}
Arc2Face~\cite{Paraperas2024Arc2Face} and face IP-Adapter variants~\cite{Zhang2023IPAdapter}
demonstrated realistic pose and expression changes without label flips.
A systematic benchmark that fixes the backbone while varying only the augmentation
generator is however missing. In this paper, we provide such an analysis and compare different augmentation generators.

\paragraph{Balanced synthetic data for fairness}
Demographic bias in FR is well documented~\cite{kotwal2024demographic,Wang2019RFW,kotwal2025review}.
Several recent studies leverage synthetic data for bias mitigation. For instance, Qraitem \emph{et al.}~\cite{Qraitem2024FTR} proposed a two-step strategy ``From Fake to Real'' (FTR) that first pre-trains on a \emph{balanced} synthetic set and then fine-tunes on real data to reduce spurious correlations in general image recognition tasks. However, FTR is agnostic to facial identities and does not study FR-specific synthetic generation or the trade-off between intra- and inter-class variations. Focusing specifically on face recognition, Melzi \emph{et al.} fine-tuned existing FR models with demographically balanced synthetic data produced by GANDiffFace and reported reductions in racial bias in RFW~\cite{Melzi2024BiasMitigation}. However, they did not equalize dataset scale when comparing to real-data baselines and did not evaluate on the challenging IJB-B/C benchmark. The authors of VariFace~\cite{yeung2025variface} claimed that they generated balanced synthetic dataset which is close in performance to the real data and they equalized all dataset sizes when comparing the performance, however, they do not show the results on IJB-B/C, and they compared with CASIA WebFace, which, despite being a real dataset, leads to poor generalization capabilities~\cite{George2025Digi2Real}.

Initiatives such as the FRSyn competition series have helped progress in the
development and use of synthetic datasets~\cite{deandres2024frcsyn,
melzi2024frcsyn1}. Despite these advancements, FR systems trained only on
synthetic data continue to exhibit inferior performance compared to those
trained on real-world datasets of similar size~\cite{melzi2024frcsyn,
George2025Digi2Real}. This performance disparity is evident not only in
recognition metrics--such as False Match Rate (FMR) and False Non-Match Rate
(FNMR)--but also in fairness across demographic groups, as indicated by
variability in group-specific outcomes. In \cite{huber2024bias}, Huber
\textit{et al.} noted that FR models trained on synthetic dataset are likely to
amplify demographic disparities, compared to the models trained on real
datasets.

\paragraph{Scale-controlled comparisons}
Performance gains reported for synthetic data are often confounded by unmatched
sample counts. Datasets such as WebFace-42M or even WebFace260M~\cite{Zhu2021WebFace260M} are much larger than most synthetic datasets and therefore always demonstrate higher performance when used to train FR models. We therefore enforce a similar number of identities \emph{and} number of images-per-identity across all training data, making our experimental results comparable in a fairer way, an approach largely missing from the existing works. Only a recent work by Yeung \emph{et al.}~\cite{yeung2025variface} compared FR models trained on the same number of IDs and images per ID. However, the paper only used standard small benchmarking datasets for evaluation (LFW~\cite{huang2008labeled}, AgeDB-30~\cite{moschoglou2017agedb}, and others), missing the results on IJB-B/C~\cite{Whitelam2017IJBB, Maze2018IJBC}, which shows an incomplete picture with respect to the generalizability of the FR models trained on synthetic data.

\section{Datasets}
\label{sec:data}

In this section, we provide details on how we have generated synthetic faces balanced by gender, race, and age that we used to train AdaFace~\cite{Kim2022AdaFace} face recognition model. We then describe the existing datasets that we use as baselines to compare our generated data for the recognition accuracy. We also describe the test datasets used to benchmark the trained FR models.

\subsection{Balanced synthetic data generation}
\label{sec:generation}
To investigate the effect of \emph{seed generator} on face recognition training, we use the prompt based technique to generate initial identities with Flux.1-Dev~\cite{yang2024flux} (denoted as Flux) and Stable Diffusion~v3.5~\cite{stabilityai2024sd3} (denoted as SD35).  Figure~\ref{subfig:a} shows an image example generated by Flux in the top row and one generated by SD35 in the bottom row. We focused on generating balanced facial data by varying each of these prompt attributes: seven age groups (below 8 years old, between ages 8 and 13, between ages 13 and 18, between ages 18 and 25, between ages 25 and 35, between ages 35 and 50 ranging, and older than 50 years), four races (Caucasian White, Black, East-Asian, and Latino Hispanic), and two genders (male and female). Given that there are 56 variations of these attributes, with 200 for each variation, we have generated a total of 11200 different faces for each generator. When generating the prompts, we also randomly varied other soft-attributes such as hairstyle, hair-color, glasses (no glasses, reading glasses or sunglasses), presence of earrings, lighting conditions, i.e., daylight, sunset, indoor lighting, low light, and raining, backgrounds, e.g., beach, street, or studio, photos styles, e.g., portrait or artistic, quality of the generated images, such as compressed, professional, or amateur; and  camera types, including professional DSLR, selfie webcam, or smartphone. Since these soft environmental attributes were chosen at random, they are equally present in all images regardless of race or age. Note that exactly the same prompts were used for both Flux and SD35, so that the differences in the resulting images are dictated solely by the differences in these text-to-image models.

To understand the impact of \emph{augmentation generator} on face recognition training, we used these publicly available models trained to generate several variations of the same input identity: Arc2Face~\cite{Paraperas2024Arc2Face} which uses face embeddings from ArcFace model~\cite{deng2019arcface} pretrained on WebFace42M to enforce ID consistency when generating the variations with Stable Diffusion v1.5 (SD15) and IP-Adapter variants~\cite{Zhang2023IPAdapter}, which use either a face embedding from InsightFace~\cite{Deng2020CVPR} or from CLIP~\cite{radford2021clip} model for ID consistency when using SD15 and Stable Diffusion XL (SDXL) backbones. Figure~\ref{fig:flux} shows SD35-based augmentations at the top and for Flux at the bottom rows. For each image generated from a \emph{seed generator}, which represents an identity, we generated the number of augmentations with these \emph{augmentation generators} ranging from 8 images per identity, to 32 images per identity with a step of 8. We vary the number of images per identity to understand whether and how this number affects the performance of face recognition. 

\subsection{State of the Art datasets}

We utilized publicly available state-of-the-art synthetic datasets such as DCFace~\cite{Kim2023DCFace}, Digi2Real~\cite{George2025Digi2Real}, and HyperFace~\cite{Otroshi2024HyperFace} as baselines to compare the performance of face recognition models trained on them. Since our synthetic datasets (see Section~\ref{sec:generation}) have about 11K identities (but this number can reduce after face detection), we have limited all baseline datasets to about 10K identities, so that the accuracy of the face recognition model trained on all the data can be fairly compared. Similarly, we have used real datasets with about 10K identities, such as widely adopted CASIA-WebFace~\cite{yi2014learning} dataset with about 10.5K identities and we took 10K identities subset from WebFace42M~\cite{Zhu2021WebFace260M}.

\begin{figure*}[htb]
\centering
\begin{subfigure}{0.4\textwidth}
\includegraphics[width=\textwidth]{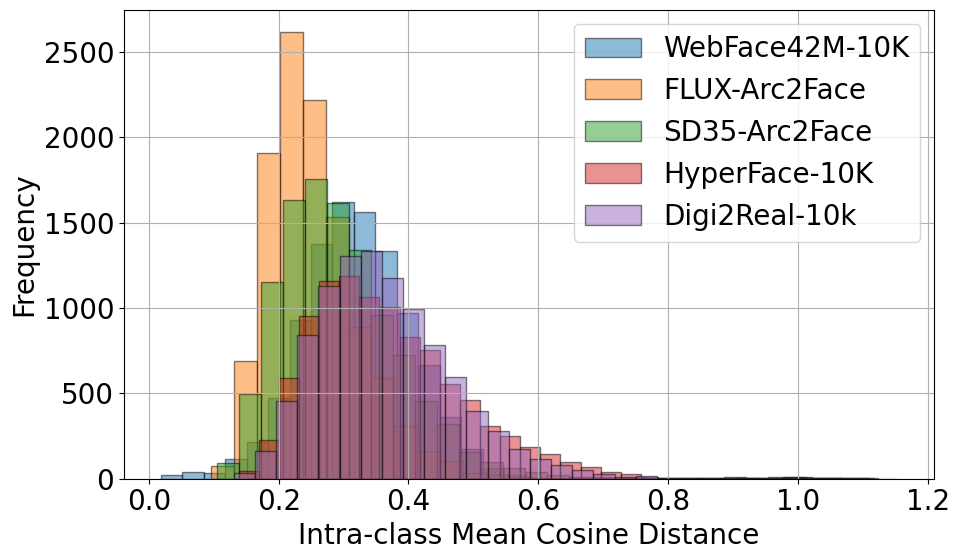}
	\caption{Mean cosine distance within the same identity}
	\label{fig:analysis_intra}
\end{subfigure}
\qquad
\begin{subfigure}{0.4\textwidth}
\includegraphics[width=\textwidth]{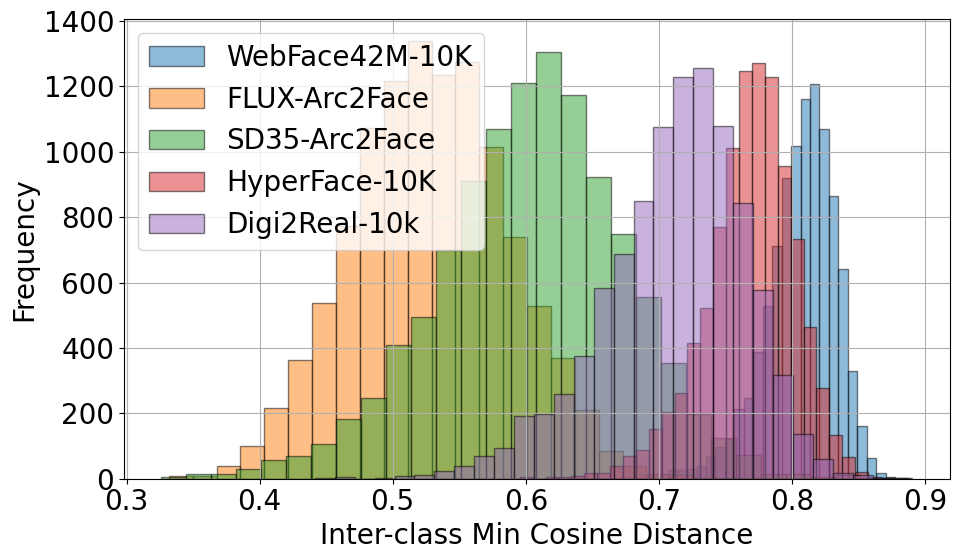}
	\caption{Min cosine between different identities}
	\label{fig:analysis_inter}
\end{subfigure}
\caption{Mean intra-class and min inter-class cosine distances for synthetic datasets and WebFace-10K.}
\label{fig:db_analysis}
\end{figure*}

\subsection{Benchmarking datasets}
\label{sec:benchmark}

We evaluate face recognition models trained on the above synthetic and real data using the \emph{standard} benchmark datasets, including Labeled Faces in the Wild (LFW)~\cite{huang2008labeled}, Cross-Age LFW (CA-LFW)~\cite{zheng2017cross}, Cross-Pose LFW (CP-LFW)~\cite{zheng2018cross}, Celebrities in Frontal-Profile in the Wild (CFP-FP)~\cite{sengupta2016frontal}, and AgeDB-30~\cite{moschoglou2017agedb}, as well as, the large public test benchmarks such as IARPA Janus Benchmark-B (IJB-B)~\cite{Whitelam2017IJBB} and IARPA Janus Benchmark-C (IJB-C)~\cite{Maze2018IJBC}. Evaluating the models on the more challenging IJB-B/C datasets is very important because the models can show very high accuracies on the \emph{standard} datasets but fail to generalize to IJB-B/C; and that is why the results on IJB-B/C are often missing from the state of the art work.

For bias assessment of trained face recognition models, we use the Racial Faces in the Wild (RFW) dataset~\cite{Wang2019RFW}, which has been widely adopted to benchmark racial bias or fairness in FR systems. It consists of
four demographic subsets-- African, Asian, Caucasian, and Indian-- each
containing approximately 3,000 unique individuals and 6,000 predefined image
pairs curated by the dataset creators. Each subset includes an equal number of
genuine (mated) and impostor (non-mated) pairs, offering a balanced and
structured framework for evaluating verification performance across racial
groups.

\subsection{Analysis of the datasets}

To analyze how well \emph{augmentation generators} preserve the identity of its different variations, i.e., their intra-class ID similarity, we compute, for the same identity, the cosine distances between the EdgeFace~\cite{george2024edgeface} embeddings of its augmentations. The mean of these cosine distances for the identity indicates how similar are the augmentations between each other. If the mean is close to zero, all augmentations/variations of this ID are similar.

Figure~\ref{fig:analysis_intra} shows the histograms of mean cosine distances computed for Arc2Face 
augmentation generator and for all identities of Flux or SD35 seed generators and compares them against HyperFace and Digi2Real synthetic datasets and real faces from the 10K IDs subset of WebFace42M. The figure shows that all datasets have similar distributions of mean cosine distances and they are as close to zero as the WebFace distribution, so the augmentation generators used in the synthetic datasets match closely the real data.

To understand how separated the identities generated by both \emph{seed generators} and \emph{augmentation generators} are from each other, i.e., their inter-class ID difference, we have computed the minimum cosine distance between the average EdgeFace embedding of a given identity with the average embedding of every other identity. Essentially, for each identity, we find the most similar looking other identity (according to EdgeFace). 
If all identities are very distinct from each other, their min cosine distance will be large. 

Figure~\ref{fig:analysis_inter} shows histograms of the min cosine distances computed for all identities of the same Flux or SD35-based seed generators, HyperFace and Digi2Real synthetic datasets, and WebFace42M real dataset. 
Figure shows that histograms of the synthetic datasets are to the left of the WebFace42M. It means the inter-class variability is lower for synthetic datasets, indicating that face recognition trained on synthetic data, potentially, should have lower discriminative power and may perform worse than the one trained on real data. The distribution of the HyperFace is the closest to that of WebFace, while the Flux-based data is the furthest away, which means that Flux seed generator may not be the best for training face recognition.

\section{Face recognition training and evaluation}
\label{sec:method}

To have a fair comparison, we have trained the same model ResNet-50 with AdaFace loss\footnote{\url{https://github.com/mk-minchul/AdaFace}} on different variations of our synthetic datasets and compared them with the baseline synthetic and real data. We evaluated all the models using the standard benchmark datasets (see Section~\ref{sec:benchmark}) and IJC-B/C datasets for general face recognition assessment and RFW dataset for bias assessments.

\renewcommand{\arraystretch}{1.2}
\begin{table*}[htb]
\setlength\tabcolsep{6pt} 
\small
\centering
\caption{Accuracy results of ResNet-50 AdaFace model trained on different real (highlighted green) and synthetic datasets. All accuracy values are reported as \%.}
\label{tab:benchmark-results}
\resizebox{\textwidth}{!}{
\begin{tabular}{lc|ccccc|cc|cc}
\toprule
 Training dataset  & \# Images & \multicolumn{5}{c|}{Standard benchmark datasets} & \multicolumn{2}{c|}{IJB-B} & \multicolumn{2}{c}{IJB-C} \\
 & per ID & AgeDB30 & CFP-FP & LFW & CP-LFW & CA-LFW & 1E-04 & 1E-03 & 1E-04 & 1E-03 \\
\midrule

\rowcolor{green!10}
WebFace42M-10K~\cite{Zhu2021WebFace260M} & 22 & 91.7 & 92.8 & 99.3 & 89.6 & 93.3 & 86.5 & 92.8 & \textbf{89.5} & \textbf{94.5} \\
\rowcolor{green!10}
CASIA WebFace~\cite{yi2014learning} & 46 & 94.7 & 95.3 & 99.5 & 90.5 & 93.8 & 13.6 & 75.3 & \textit{15.9} & \textit{73.8} \\
\rowcolor{red!10}
DCFace-10K-Small & 20 & 88.4 & 83.7 & 98.0 & 79.6 & 90.8 & 0.1 & 0.8 & \textit{0.1} & \textit{0.6} \\
\rowcolor{red!10}
DCFace-10K~\cite{Kim2023DCFace} & 54 & 90.3 & 88.3 & 98.3 & 83.5 & 91.7 & 77.0 & 86.3 & \textbf{81.0} & \textbf{89.2} \\
\rowcolor{red!10}
Digi2Real-10K~\cite{George2025Digi2Real} & 20 & 82.6 & 86.3 & 98.5 & 81.5 & 87.4 & 67.5 & 80.0 & 73.0 & 83.5 \\
\rowcolor{red!10}
HyperFace-10K~\cite{Otroshi2024HyperFace} & 64 & 87.1 & 89.1 & 98.7 & 84.7 & 89.8 & 67.3 & 84.7 & 75.7 & 87.3 \\
\rowcolor{orange!10}
SD35-All-IPA & 33 & 79.4 & 80.3 & 94.8 & 75.7 & 83.2 & 54.8 & 68.9 & 58.6 & 72.2 \\
\rowcolor{orange!10}
SD35-Arc2Face-All-IPA & 41 & 79.4 & 81.0 & 95.2 & 75.7 & 83.4 & 55.7 & 69.3 & 59.0 & 72.6 \\
\rowcolor{orange!10}
SD35-Arc2Face-8 & 8 & 77.0 & 75.3 & 94.1 & 72.6 & 81.5 & 56.7 & 68.3 & 60.6 & 71.8 \\
\rowcolor{orange!10}
SD35-Arc2Face-16 & 16 & 82.2 & 83.2 & 96.6 & 77.4 & 85.4 & 64.9 & 75.8 & \textbf{68.9} & \textbf{78.8} \\
\rowcolor{orange!10}
SD35-Arc2Face-24 & 24 & 83.1 & 83.7 & 97.0 & 78.9 & 86.2 & 19.6 & 68.7 & 27.0 & 72.5 \\
\rowcolor{orange!10}
SD35-Arc2Face-32 & 32 & 83.9 & 85.3 & 97.2 & 78.9 & 85.6 & 6.6 & 51.4 & \textit{9.0} & \textit{57.8} \\
\rowcolor{cyan!10}
Flux-All-IPA & 33 & 78.4 & 79.9 & 95.1 & 74.4 & 83.6 & 55.4 & 66.5 & 59.0 & 69.9 \\
\rowcolor{cyan!10}
Flux-Arc2Face-All-IPA & 41 & 80.8 & 77.1 & 96.0 & 73.9 & 85.6 & 61.3 & 73.0 & 64.2 & 76.1 \\
\rowcolor{cyan!10}
Flux-Arc2Face-8 & 8 & 73.0 & 73.1 & 93.5 & 71.7 & 81.3 & 53.8 & 66.4 & 57.8 & 70.0 \\
\rowcolor{cyan!10}
Flux-Arc2Face-16 & 16 & 80.1 & 78.9 & 96.2 & 75.4 & 85.3 & 65.0 & 75.7 & 69.2 & 78.7 \\
\rowcolor{cyan!10}
Flux-Arc2Face-24 & 24 & 80.6 & 80.4 & 97.0 & 76.7 & 86.5 & 65.0 & 76.0 & \textbf{68.8} & \textbf{79.2} \\
\rowcolor{cyan!10}
Flux-Arc2Face-32 & 32 & 80.9 & 82.1 & 96.9 & 77.2 & 86.5 & 18.4 & 58.3 & \textit{19.9} & \textit{61.3} \\

\bottomrule
\end{tabular}
}
\end{table*}
 
\renewcommand{\arraystretch}{1.2}
\begin{table*}[ht]
\centering
\caption{Bias assessment on RFW of ResNet-50 AdaFace model trained on different real (highlighted green) and synthetic datasets. Accuracy values are reported as \%.}
\label{tab:bias_results}
\begin{tabular}{lrrrrrrr}
\toprule
Training dataset    &
\multicolumn{4}{c}{Group-Level Accuracy \color{green}{$\uparrow$}} & 
Avg Acc \color{green}{$\uparrow$}   & Std  \color{red}{$\downarrow$}&
TO  \color{green}{$\uparrow$}  \\ \cline{2-5}
    & African  & Asian  & Caucasian  & Indian  &  &  &  \\ \midrule

\rowcolor{green!10} 
WebFace42M-10K~\cite{Zhu2021WebFace260M}  & 81.80 & 84.20 & 90.12 & 85.12 & 85.31 & 3.03 & 82.28 \\
\rowcolor{green!10} 
CASIA WebFace~\cite{yi2014learning}  & 87.40 & 85.07 & 92.13 & 87.60 & 88.05 & 2.56 & \textbf{85.49} \\
\rowcolor{red!10} 
DCFace-10K-Small~\cite{Kim2023DCFace} & 69.10 & 77.28 & 83.33 & 79.05 & 77.19 & 5.16 & 72.03 \\
\rowcolor{red!10} DCFace-10K~\cite{Kim2023DCFace} & 73.33 & 79.38 & 86.22 & 82.00 & 80.23 & 4.67 & 75.56 \\
\rowcolor{red!10} Digi2Real~\cite{George2025Digi2Real} & 77.52 & 76.82 & 82.97 & 78.63 & 78.98 & 2.39 & 76.59 \\
\rowcolor{red!10} HyperFace~\cite{Otroshi2024HyperFace} & 74.88 & 81.50 & 86.27 & 83.32 & 81.49 & 4.18 & 77.31 \\
\rowcolor{orange!10} SD35-All-IPA & 74.60 & 74.58 & 77.18 & 75.43 & 75.45 & \textbf{1.06} & 74.39 \\
\rowcolor{orange!10} SD35-Arc2Face-All-IPA & 73.55 & 74.85 & 76.82 & 74.72 & 74.98 & 1.17 & 73.81 \\
\rowcolor{orange!10} SD35-Arc2Face-16 & 76.45 & 76.55 & 79.43 & 76.73 & 77.29 & 1.24 & 76.05 \\
\rowcolor{orange!10} SD35-Arc2Face-24 & 78.17 & 78.45 & 81.42 & 77.73 & 78.94 & 1.45 & \textbf{77.49} \\
\rowcolor{cyan!10} Flux-All-IPA & 68.25 & 71.45 & 74.12 & 69.25 & 70.77 & 2.25 & 68.51 \\
\rowcolor{cyan!10} Flux-Arc2Face-All-IPA & 73.15 & 75.50 & 77.77 & 71.28 & 74.42 & 2.44 & 71.98 \\
\rowcolor{cyan!10} Flux-Arc2Face-16 & 74.78 & 76.32 & 77.90 & 73.90 & 75.72 & 1.52 & 74.20 \\
\rowcolor{cyan!10} Flux-Arc2Face-24 & 76.27 & 76.78 & 79.20 & 75.12 & 76.84 & 1.49 & 75.35 \\
\bottomrule
\end{tabular}
\end{table*}

\subsection{Face recognition setup} 
\label{sec:recognition}

For all experiments 
we followed a consistent FR pipeline. Each input image underwent standardized
preprocessing that included face detection and facial landmark localization
using MTCNN~\cite{zhang2016joint}, followed by
geometric alignment and resizing of the cropped face region to a resolution of
$112 \times 112$ pixels in RGB format that conforms to the input requirements of
the ResNet-50 architecture trained using AdaFace loss~\cite{Kim2022AdaFace}. On all datasets, the training was done for $30$ epochs with batch size $256$ and an initial learning rate of
$0.1$, which was reduced by a factor of 10 at epochs 12, 20, and 26. We also applied with $0.2$ probability the random data augmentations provided in the AdaFace code\footnotemark[2] by default: central cropping, resizing, horizontal flip, and photometric augmentation. At inference, the resulted model generates a $512$-$d$ feature
representation (\textit{embedding}) for each input presentation. These
embeddings were subsequently compared using cosine similarity to perform
identity matching.

\subsection{Evaluation metrics} 
\label{sec:metrics}

We evaluated the overall recognition performance of the FR models using accuracy metrics on the benchmarking datasets (see Section~\ref{sec:benchmark}), as it is typically done in the current literature on face recognition~\cite{kotwal2024demographic, melzi2024frcsyn1, kotwal2024mitigating}. For IJB-B/C datasets~\cite{Whitelam2017IJBB,Maze2018IJBC}, we report  the True Accept Rate (TAR) at a False Accept Rate (FAR) of 1E-3 and  1E-4. We use RFW dataset~\cite{Wang2019RFW} for the analysis of bias across racial
demographics, measuring bias as the standard
deviation of accuracy across its demographic subgroups. For all experiments, we
first determined the optimal decision threshold-- yielding the highest possible
accuracy -- on the complete RFW dataset, and then applied this same threshold
across each racial subgroup to compute group-specific accuracy values.

\begin{figure*}[htb]
\centering
\includegraphics[width=0.4\textwidth]{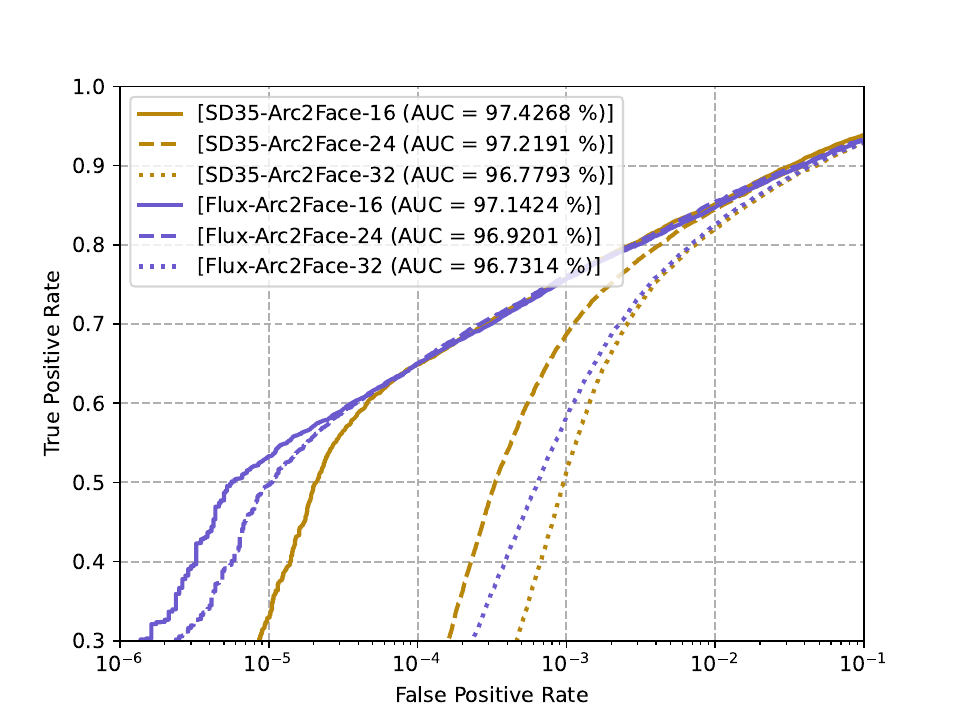}
\qquad
\includegraphics[width=0.4\textwidth]{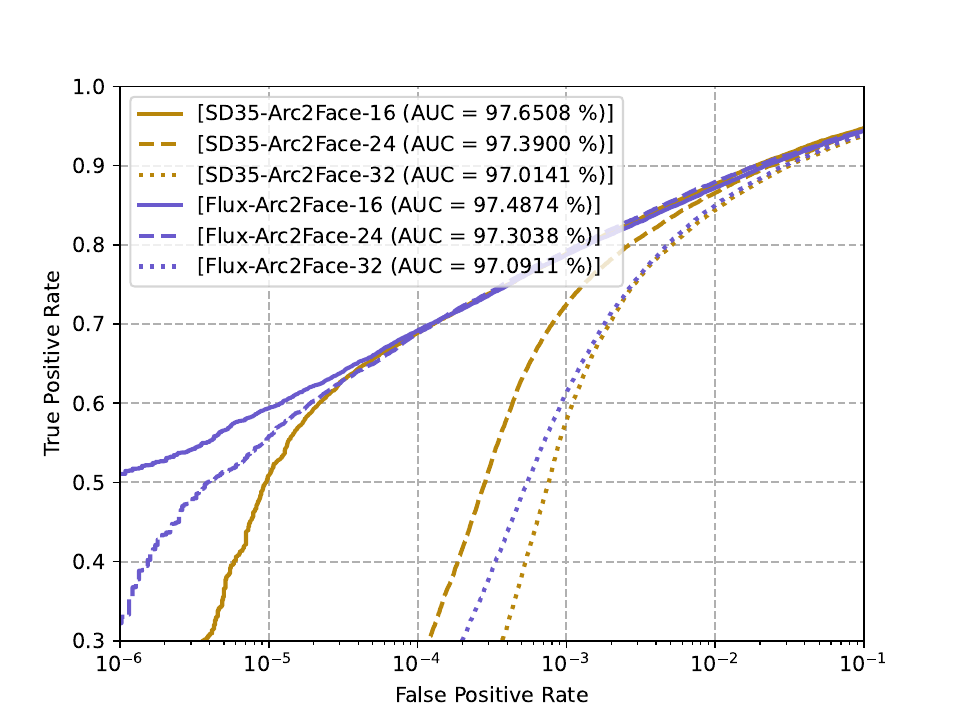}
\caption{ROC computed on IJB-B and IJB-C databases for AdaFace models trained on faces generated with 16, 24, or 32 images per ID by Arc2Face augmentation generator from Flux (blue color) or SD35 (golden color) seed identities.}
\label{fig:roc}
\end{figure*}

Given the necessity to assess both accuracy and fairness in face recognition models, we also
used the trade-off (TO) metric, which has been used in recent studies on
synthetic data in FR~\cite{deandres2024frcsyn, melzi2024frcsyn1}. Simply put, TO metric is the difference between an average accuracy and the standard deviation of accuracy across demographic groups.

\section{Experimental results}
\label{sec:results}

In this section, ResNet-50 AdaFace model\footnotemark[2], trained on different real and synthetic datasets, is evaluated on standard benchmark datasets for face recognition and on a balanced RFW dataset for bias assessment.

\subsection{Benchmarking on standard datasets}
Table~\ref{tab:benchmark-results} shows the performance of ResNet-50 AdaFace model\footnotemark[2] trained on different synthetic and real datasets and evaluated on standard benchmarking and IJB-B/C datasets (see Section~\ref{sec:benchmark} for details). All datasets in the table have about 10-11K identities but the number of images per ID can vary significantly and is indicated in the `\#Images per ID' column. In the table, real datasets are colored green, baseline synthetic dataset are colored red, the data generated from SD35 is orange, and from Flux is blue. For each category/color, the best performing on IJB-B/C dataset is set in \textbf{bold} and the worst is in \textit{italic}.

Table~\ref{tab:benchmark-results} highlights several important issues:

\begin{itemize}
    \item High accuracy on the five standard benchmark datasets (AgeDB30, CFP-FP, etc.) does not always translate to the high true accept rate (TAR) on IJB-B/C datasets. A notable example is CASIA WebFace showing the highest accuracy on the five benchmark datasets with an average at $94.8$\% but very low performance on both IJB-B/C datasets. The overall size of CASIA WebFace is third largest with $46$ images per ID (It is an average number for real datasets), only full DCFace 0.5M (marked as `DCFace-10K' in the table) and HyperFace have more total images, however, it did not help it to generalize to IJB-B/C datasets.
    \item A 10K subset of real faces from WebFace42M leads to the best overall performance. The closest synthetic dataset in performance is DCFace-10K with an average accuracy of $90.4$\% on five standard benchmarks vs $93.4$\% for WebFace42M and with about $10$\% smaller in TAR on IJB-B/C. 
    \item In synthetic datasets, the number of images per ID has a significant impact on the performance. If one looks at five benchmark datasets only, for the same generator, the more images per ID lead to better accuracy. However, when looking at IJB-B/C results, the performance can be very different. Notably, DCFace-10K-Small, which is obtained from the actual DCFace 0.5 (marked as `DCFace-10K' in the table) by randomly sub-sampling the number of images per ID from $54$ to $20$, has almost zero true positive rates on IJB-B/C. The same identities, just about half of images per ID kept, and the performance drops from the highest for synthetic datasets (DCFace-10K) to zero (DCFace-10K-Small) on IJB-B/C. This surprising result illustrates that a dedicated evaluation of the DCFace would be required to better understand this behavior, but such investigation is out of the scope of this paper. 
    \item The performance for Flux-based generated images, for the same parameters, is slightly higher on IJB-B/C than for SD35-based images. In terms of the augmentation generator, when mixing all augmentation generators, Arc2Face and all four IP-Adapters (denoted by `All-IPA') and original seed images, it leads to the drop in performance on IJB-B/C. This is probably due to too much intra-class variation inside each ID (see examples in Figure~\ref{fig:flux}), with face augmentations of one ID possibly leaking into other IDs. 
    \item We evaluate the direct impact of the number of augmentations per ID by fixing all parameters (for generation and training) except for the number of images generated for each ID by Arc2Face augmentation generator. From the results for both Flux and SD35, it is clear that when we increase the number of images per ID from $8$ to $16$, the performance show a clear relative improvement on all benchmarks. When we increase the the number of images per ID to $24$, and then to $32$, the accuracy on five benchmarks improves on average. However, it is different for IJB-B/C evaluations. For SD35-based data, the performance on IJB-B/C drops, once the number of images per ID become $24$ and drops even more when the number is $32$. For Flux-based images, for both $16$ and $24$ images per ID, the performance on IJB-B/C is similar and drops only when the number of images is $32$. To demonstrate this phenomena in more details, we have plotted the receiver operating characteristic (ROC) curves for $16$, $24$, and $32$ images per ID for SD35 and Flux-based data in Figures~\ref{fig:roc}. The figures show ROCs moving to the right with each increase of number of images from 16, 24, and 32, for either SD35 or Flux based data.
\end{itemize}

\subsection{Bias assessment}

Table~\ref{tab:bias_results} shows the bias assessment results (on RFW~\cite{Wang2019RFW} race-balanced dataset) of ResNet-50 AdaFace model\footnotemark[2] trained on different synthetic and real data. The main observation from `Std' column, which essentially shows the bias of the model (the lower value, the lesser the bias) to the faces of different races, is that SD35-based synthetic data have the lowest overall standard deviation with the lowest value of $1.06$ for SD35-All-IPA where all IP-Adapters are equally mixed together as augmentations of each identity. Even real datasets have higher standard deviation values. Clearly, the average accuracy of SD35-based synthetic data is lower than for real datasets or the synthetic HyperFace dataset (though its Std measuring bias is very high). High accuracy of HyperFace on RFW and IJB-B/C can be explained by its higher inter-class variability, which closely matches that of real data of WebFace42M, as illustrated by Figure~\ref{fig:analysis_inter}. Hence, if we make inter-class variability of SD35 higher, we may improve its overall face recognition performance, while still mitigating the bias better than others thanks to its race-balanced nature. 

Overall, synthetic datasets generated without balancing the data exhibit high bias, with an exception of Dig2Real, which is on par with some of the Flux-based data. 
For real datasets, CASIA WebFace outperforms the 10K subset from WebFace42M, however, we should not forget that CASIA WebFace has a very poor performance on a more challenging IJB-B/C benchmark (see Table~\ref{tab:benchmark-results}). 

If we consider a one-stop `TO' metric~\cite{deandres2024frcsyn, melzi2024frcsyn1}, computed as a difference between average accuracy and standard deviation, SD35-Arc2Face-24 has the highest TO value among the synthetic datasets. However, it is important to remember that this dataset led to very poor performance on IJB-B/C in Table~\ref{tab:benchmark-results}. It means that in practical applications, one needs to consider and conduct the evaluations of both bias and challenging face recognition benchmarks to find the tradeoff between an overall accuracy and bias mitigation properties of the synthetic data used for training.

From Table~\ref{tab:bias_results}, we can note that SD35-based data is more suitable for bias mitigation. A possible reason is that the identities distribution of SD35 is more similar to the real faces compared to Flux as shown in Figure~\ref{fig:analysis_inter}. Another reason is the high variations in visual appearance of images generated by SD35 compared to Flux. Visually, the Flux generates more professionally looking portraits that are \emph{visually appealing} but probably not suitable for training a fairer face recognition. SD35 images look more like images collected in the wild. Overall, the SD35-based images are suited more for bias mitigation, especially, if they can be combined with real images, thus, improving the overall accuracy while keeping the bias low.


\section{Conclusion}

This paper presents a comprehensive study of how balanced synthetic datasets affect both the accuracy and bias of face recognition systems. Through controlled experiments using text-to-image models and identity-consistent augmentation methods, we demonstrate that demographic balancing during synthetic data generation can significantly reduce racial bias in FR models, as evidenced by lower standard deviation in group accuracies on RFW. However, synthetic data alone still underperforms compared to real datasets on challenging benchmarks like IJB-B/C, particularly when seed identities are less separated from each other. In bias evaluations, synthetic data generated using SD35 consistently achieved better fairness metrics than other methods, including the real datasets, suggesting that using synthetic balanced datasets could help for training fairer practical face recognition systems.
Our findings highlight the importance of thoughtful design choices in seed and augmentation generators, and indicate that hybrid training approaches, combining synthetic and real data, may be a promising path forward to achieve both high performance and fairness in face recognition systems.


\section*{Acknowledgement}
This work was funded by InnoSuisse 106.729 IP-ICT.

\balance
{\small
\bibliographystyle{ieeetr}
\bibliography{references}
}

\end{document}